\documentclass[journal]{IEEEtran}
\usepackage{cite}

\ifCLASSINFOpdf
\usepackage[pdftex]{graphicx}
\graphicspath{{./figure/}}
\else
\usepackage[dvips]{graphicx}
\graphicspath{{./figure/}}
\fi
\usepackage{amsmath}
\usepackage{mathtools}

\usepackage{tabularx,ragged2e}
\newcolumntype{C}{>{\Centering\arraybackslash}X} 

\usepackage{xcolor}
\definecolor{applegreen}{rgb}{0.55, 0.71, 0.0}

\ifCLASSOPTIONcompsoc
 \usepackage[caption=false,font=normalsize,labelfont=sf,textfont=sf]{subfig}
\else
 \usepackage[caption=false,font=footnotesize]{subfig}
\fi
\usepackage{url}

\begin{document}
%

\title{Using Reinforcement Learning with Partial Vehicle Detection for Intelligent Traffic Signal Control}
%
%
%

\author{\IEEEauthorblockN{R.~Zhang \IEEEauthorrefmark{1}, A.~Ishikawa \IEEEauthorrefmark{1}, W.~Wang \IEEEauthorrefmark{1}, B.~Striner \IEEEauthorrefmark{2}, and O.K.~Tonguz \IEEEauthorrefmark{1}}\\
\IEEEauthorblockA{\IEEEauthorrefmark{1} Department of Electrical and Computer Engineering,
Carnegie Mellon University,
Pittsburgh, PA 15213-3890,
USA
}\\
\IEEEauthorblockA{\IEEEauthorrefmark{2} Machine Learning Department,
Carnegie Mellon University,
Pittsburgh, PA 15213-3890,
USA
}
}
\maketitle


\begin{abstract}
Intelligent Traffic Signal Control (ITSC) systems have attracted the attention of researchers and the general public alike as a means of alleviating traffic congestion. Recently, the vehicular wireless technologies have enabled a cost-efficient way to achieve ITSC by detecting vehicles using Vehicle to Infrastructure (V2I) wireless communications.

Traditional ITSC algorithms, in most cases, assume that every vehicle is detected, such as by a camera or a loop detector, but a V2I implementation would detect only those vehicles equipped with wireless communications capability. We examine a family of transportation systems, which we will refer to as `Partially Detected Intelligent Transportation Systems'. An algorithm that can perform well under a small detection rate is highly desirable due to gradual increasing penetration rates of the underlying technologies such as Dedicated Short Range Communications (DSRC) technology. 
Reinforcement Learning (RL) approach in Artificial Intelligence (AI) could provide indispensable tools for such problems where only a small portion of vehicles are detected by the ITSC system.
 


In this paper, we report a new RL algorithm for Partially Detected Intelligent Traffic Signal Control (PD-ITSC) systems. The performance of this system is studied under different car flows, detection rates, and typologies of the road network. Our system is able to efficiently reduce the average waiting time of vehicles at an intersection, even with a low detection rate, thus reducing the travel time of vehicles.

\end{abstract}


%
\IEEEpeerreviewmaketitle

\section{Introduction}
\footnote[1]{The research reported in this paper was partially funded by King Abdulaziz City of Science and Technology (KACST), Riyadh, Kingdom of Saudi Arabia}
Traffic congestion is a daunting problem that affects the daily lives of billions of people in most countries across the world \cite{congestion}. Over the last 30 years, many Intelligent Traffic Signal Control (ITSC) systems have been designed and demonstrated as one of the effective way to reduce traffic congestion \cite{robertson1969tansyt, lowrie1990scats, hunt1982scoot,luk1984two,gartner1983opac,mirchandani2001real,henry1984prodyn,vincent1988mova}. These systems use real time traffic information measured or collected by video cameras or loop detectors  and optimize the cycle split of a traffic light accordingly \cite{trafficLight}. Unfortunately, such intelligent traffic signal control schemes are expensive and, therefore, they exist only at a small percentage of intersections in the United States, Europe, and Asia.


Recently, several more cost-effective approaches to implement ITSC systems were proposed by leveraging the fact that Dedicated Short-Range Communication (DSRC) technology \cite{ferreira2010self, nafi2012vanet,milanes2012intelligent}.
DSRC technology is potentially a much cheaper technology for detecting the presence of vehicles on the approaches of an intersection. However, at the early stages of deployment, only a small percentage of vehicles will be equipped with DSRC radios.  Meanwhile, the rapid development of the Internet of Things (IoT) has created new technology applicable for sensing vehicles for ITSC. Other than DSRC, applicable technologies include, but are not limited to, RFID, Bluetooth, Ultra-Wide Band (UWB), Zigbee, and even cellphone apps such as Google Map \cite{chattaraj2009intelligent, friesen2015bluetooth, qu2010intelligent}. All these systems are more economical than traditional loop detectors or cameras. Performance-wise, most of these systems are able to track vehicles in a continuous manner, while loop detectors can only detect the presence of vehicles. These ITSC systems mentioned above are all promising technologies that could bring the expensive price of traditional ITSC systems down dramatically; however, these systems have a common critical shortcoming: they are not able to detect vehicles unequipped with the communication device (i.e., DSRC radios, RFID tags, Bluetooth device, etc.). 

Since this adoption stage of the aforementioned systems could possibly take several years \cite{average}, new control algorithms that can handle partial detection of vehicles are required. One potential AI algorithm that could be very helpful is deep reinforcement learning (DRL), which has recently been explored by several groups \cite{genders2016using,van2016deep}. These results show an improvement in terms of waiting time and queue length experienced at an intersection in a fully-observable environment. Hence, in this paper, we investigate this promising approach in a \textbf{ partially observable environment}. As expected, we observe an asymptotically improving result as we increase the penetration rate of DSRC-equipped vehicles.


In this paper, we explore the capability of DRL for handling ITSC systems using partial detection.  For simplicity, in some sections, we use DSRC detection based system as the example system, but the scheme described in this paper is very general and therefore can be used for any possible forms of partial detection, such as vehicle detection based on RFID, Bluetooth Low Energy 5.0 (BLE 5.0), cellular (LTE or 5G). 
Via extensive simulations we analyze the performance of the Reinforcement Learning (RL) method.
Our results clearly show that  reinforcement learning is capable of providing an excellent traffic management scheme that is able to reduce the waiting time of commuters at intersections, even at a low penetration rate. The results also show a different performance in detected vehicles and undetected vehicles, suggesting a built-in business model, which could be the key to eventually push forward on large-scale deployment of ITSC.

The remainder of this paper is organized as follows. In Section \ref{section:related}, we review the related work in this area. Section \ref{section:PF} gives a detailed Problem formulation. Section \ref{section:aproach} outlines the Approach we use.  Section \ref{section:performance} presents the Results of our study  in terms of performance and sensitivity to critical system parameters. In Section \ref{ss:discussion}, a Discussion is presented that highlights the practical implications of our results for intelligent transportation systems in addition to highlighting important extensions of our work for future work. Finally, Section \ref{section:conclusion} concludes the paper.

\section{Related Works}
\label{section:related}
Traffic signal control using Artificial Intelligence (AI), especially reinforcement learning (RL), has been an active field of research for the last 20 years. In 1994, Mikami, et al. proposed distributed reinforcement learning (Q-learning) using a Genetic Algorithm to present a traffic signal control scheme that effectively increased throughput of a road network \cite{mikami1994genetic}. Due to the limitations of computing power in 1994, however, it could not be implemented at that time.

Bingham proposed RL for parameter search of a fuzzy-neural traffic signal controller for a single intersection \cite{bingham2001reinforcement}, while Choy et al. adapted RL on the fuzzy-neural system in a cooperative scheme, achieving adaptive control for a large area \cite{choy2002hybrid}. These algorithms are based on RL, but the major goal of RL is \textit{parameter tuning of the fuzzy-neural system}. Abdulhai et al. proposed the first truly adaptive traffic signal which learns to control the traffic signal dynamically based on a Cerebellar Model Articulation Controller (CMAC), as a Q-estimation network \cite{abdulhai2003reinforcement}.  Silva, et.al. and Oliveira et.al. then proposed a context-detector (CD) in conjunction with RL to further improve the performance under non-stationary traffic conditions \cite{da2006adaptive,de2006reinforcement}. Several researchers have focused on multi-agent reinforcement learning for implementing it on a large scale \cite{abdoos2011traffic, medina2012traffic, el2013multiagent, khamis2014adaptive}. 

Recently, with the development of GPU and computing power, DRL has become an attractive method in several fields. Several attempts have been made using Deep Q-learning for ITSC system, including \cite{li2016traffic,genders2016using,van2016deep,van2016video}. These results show that a DQN based Q-learning algorithm is capable of optimizing the traffic in an intelligent manner. 

Traditional intelligent traffic signal systems use loop detectors, magnetic detectors and cameras for improving the performance of traffic lights. In the past few decades, various adaptive traffic systems were developed and implemented.  Some of these traffic systems such as SCOOT \cite{hunt1982scoot}, SCATS \cite{lowrie1990scats}, are based on dynamic traffic coordination \cite{ luk1984two}, and can be viewed as a traffic-responsive version of TRANSYT \cite{robertson1969tansyt}. These systems optimize the offsets of traffic signals in the network, based on current traffic demand, and generate `green-wave' for major car flow.  Meanwhile, some other model-based systems have been proposed, including OPAC \cite{gartner1983opac}, RHODES\cite{mirchandani2001real}, PRODYN\cite{henry1984prodyn}. These systems use both the current traffic arrivals and the prediction of future arrivals, and choose a signal phase planning that which optimize the objective functions. While these systems work efficiently, they do have some significant shortcomings. The cost of these systems are generally very high \cite{cost, SCATScost}. 

Even though RL yields impressive results for these cases, it does not outperform current systems. Hence, the progress of these algorithms, while interesting, is of limited impact, since traditional ITSC systems perform comparably.

Meanwhile, the recent advancements in Vehicle-to-Everything (V2X) communication have made traffic signal control schemes based on such technology a rising field, as the cost is significantly lower than a traditional ITSC system \cite{ferreira2010self, nafi2012vanet,milanes2012intelligent}. Within these schemes, a system known as Virtual Traffic Lights (VTL) is very attractive, as it proposes an infrastructure-free DSRC-based solution, by installing traffic control devices in vehicles and having the vehicles decide the right-of-way at an intersection locally. Different aspects of VTL technology have been studied by different research groups in the last few years \cite{ferreira2010self,neudecker2012feasibility,ferreira2012impact,nakamurakare2013prototype,viriyasitavat2013accelerating,bazzi2014distributed,hagenauer2014advanced,tonguz2014implementing,yapp2015safety,bazzi2016distributed,tonguz2016self,zhang2018virtual, tonguzred}.  However, a VTL system requires all vehicles in the road network to be equipped with DSRC devices, therefore, a transition scheme for the current transportation systems to smoothly transition to VTL system is needed. 

On the other hand, several methods have been proposed for floating vehicle data gathered from Global Position System (GPS) that are used to detect, estimate and predict traffic states based on fuzzy logic, Genetic Algorithm (GA), Support Vector Mechine (SVM) and other statistical learning algorithms \cite{lu2003congestion, kerner2005traffic, pattara2006estimating, de2008traffic, feng2014probe, kong2016urban}. The success of these works suggest the possibility to optimize traffic control based on partial detection (such a system is formally introduced in section \ref{section:PF}). 

There are a few research projects currently available using partial detection. For example, COLOMBO is one of the projects that focuses on low-penetration rate of DSRC-equipped vehicles \cite{bellavista2014implementing,krajzewicz2013colombo,bellavista2014v2x}.  The system uses information provided by V2X technology and feed the information to a traffic management system. Since COLOMBO cannot directly react to real-time traffic flow (the detected and undetected vehicles have the same performance), under low to medium car flow it will NOT achieve optimum performance as the optimal strategy under low-to-medium car flow has to react according to detected car arrivals.  Another very recent system is DSRC-Actuated Traffic Lights, which is one of our previous implementations using DSRC radio for traffic control. The designed prototype of this system was publicly demonstrated in Riyadh, Saudi Arabia, in July 2018 \cite{zhang2018increasing, tonguz2019harnessing}. DSRC-Actuated Traffic Lights, however, is based on the arrival of each vehicle, and hence works well under low to medium car flow rates, but it does not work well under high car flow rate. 

The main contributions of this paper are:
\begin{enumerate}
\item Explore a new kind of intelligent system that is based on \textbf{partial detection} of vehicles, which is a cost-effective alternative to current ITSC systems and an important problem not addressed by traditional ITSC systems.

\item Propose a transition scheme to VTL. Not only do we reduce the average commute time for all users, but those users that can be detected have much lower commute time, which attracts additional users to adopt the device or service. 

\item Design a new RL-based traffic signal control algorithm and system design that performs well under low penetration ratio and detection rates.

\item Provide a detailed performance analysis. Our results show that, under a low detection rate, the system can perform almost as good as an ITSC system that employs full detection. This is a very attractive solution considering its cost-effectiveness. 
\end{enumerate}

\section{Problem Statement}
\label{section:PF}

\subsection{What is a Partial Detection based ITSC System ?}
\begin{figure}[h]
\centering
\includegraphics[width=3in]{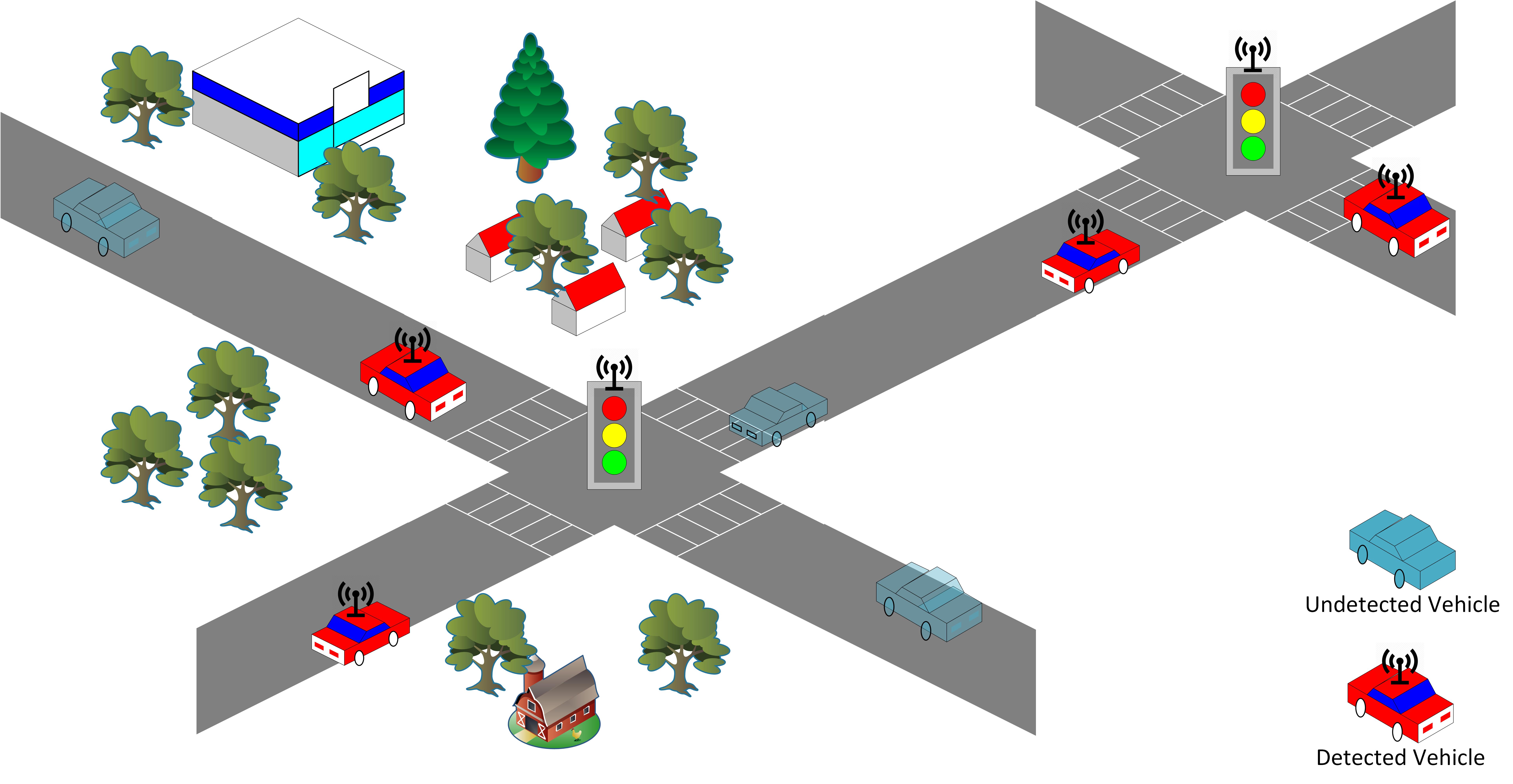}
\caption{Illustration of Partially Detected Intelligent Transportation System}
\label{fig_PDTS}
\end{figure}


Figure.\ref{fig_PDTS} gives an illustration of a \emph{Partially Detected Intelligent Traffic Signal Control} (PD-ITSC) system. There are two kinds of vehicles in the system: the red vehicles in the figure are the vehicles that the traffic lights are able to detect, we denote these vehicles as \emph{detected vehicles}; the blue semi-transparent vehicles in the figure, on the other hand, are undetectable by the traffic system, are denoted as \emph{undetected vehicles}.  In a PD-ITSC system, both kinds of vehicles co-exist in the system. The system, based on the information from the detected vehicles, decide the current phase at the intersections, in order to minimize the delay at the intersection for both detected vehicles and undetected vehicles.

Many example systems can be categorized as PD-ITSC, especially the newly proposed systems from the last decade based on wireless communications and IoT \cite{chattaraj2009intelligent, friesen2015bluetooth, qu2010intelligent}. In these systems, the vehicles are equipped with communication devices that communicate with traffic lights. Vehicles equipped with the communication device are detected vehicles and vehicles NOT equipped with the device are undetected vehicles.

In this paper, we choose one of the typical PD-ITSC system, the traffic signal system based on DSRC radios, as an example. The detected vehicles are vehicles equipped with DSRC radios, and the undetected vehicles are those unequipped with DSRC radios. Observe  that other kinds of PD-ITSC system are analogous, thus making the methodologies described in this paper applicable to them as well.

\subsection{Example PD-ITSC System Design based on DSRC}
\label{SS:ExampleSystem}

\begin{figure}[ht]
\centering
\includegraphics[width=3in]{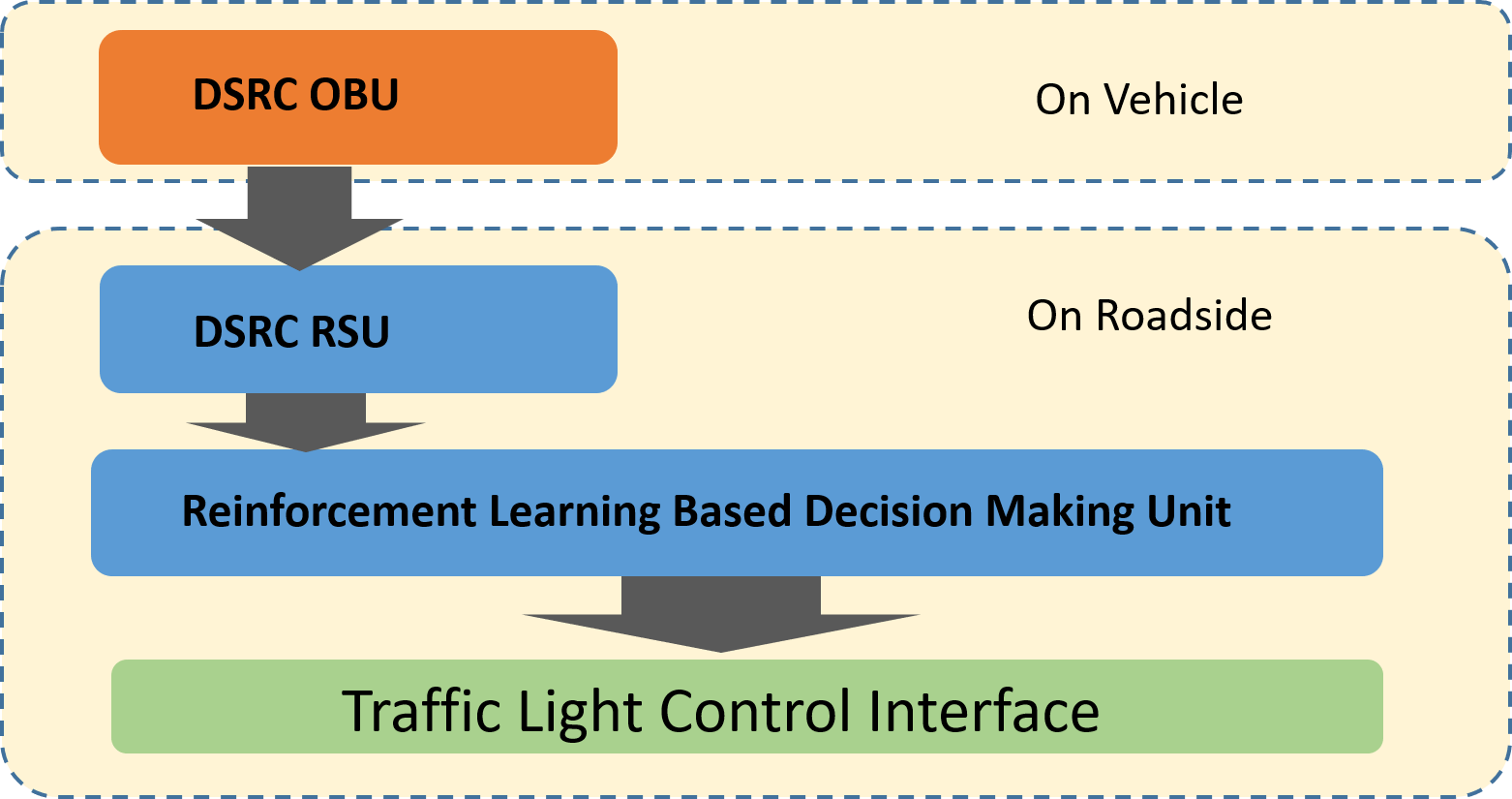}
\caption{One possible system design for the proposed scheme}
\label{fig_system}
\end{figure}

We provide here one of the possible system realizations for the proposed scheme, based on Dedicated Short-Range Communications (DSRC). The system has an 'On Roadside' unit and an 'On Vehicle' unit, as shown in Figure \ref{fig_system}. DSRC RoadSide Unit(RSU) senses the Basic Safety Message (BSM) broadcast by the DSRC OnBoard Unit (OBU), parse the useful information out, and send them to the RL Based Decision Making Unit. This unit will then make a decision based on the information provided by the RSU.


Even though the example system won't be able to detect all vehicles, it will collect more detailed information about the detected vehicles: While in traditional ITSC systems based on loop detectors, only the vehicle occupancy is detected, the system based on DSRC technology can provide a rich set of attributes including speed, distance, trajectory, and even destination of each detected vehicle. It is worth mentioning here that such properties are NOT unique to the example system considered in this section that uses DSRC technology; in fact, the same properties exist in most of other partial detection ITSC systems as well since they are based on similar wireless technologies. Therefore, the algorithm designed for PD-ITSC handling the PD-ITSC systems should be able to integrate all these pieces of information. Obviously, developing a pure analytical algorithm that takes all these information into consideration is non-trivial, thus making RL a very attractive and promising method, as it does not require a comprehensive theoretical analysis of the environment to find a near-optimal solution.

It is clear that since most of the traditional ITSC schemes do not take undetected vehicles into account, they are not suitable for PD-ITSC systems. Moreover, an ideal scheme for PD-ITSC should also:
\begin{enumerate}
\item perform well even with a low detection rate;
\item accelerate the transition to a higher adoption rate and therefore a higher detection rate (this point will be discussed in more details in Section \ref{ss:discussion}).
\end{enumerate}

\section{Approach and the Underlying Theory}
\label{section:aproach}
\subsection{Q-Learning Algorithm}
We refer to Watkins \cite{watkins1992q} for a detailed explanation of  general reinforcement learning and Q-learning but we will provide a brief review of the underlying theory in this section.

The goal of reinforcement learning is to train an agent that interacts with the environment by selecting the action in a way that maximizes the future reward. At every time step, the agent gets the state (the current observation of the environment) and reward information (the quantified indicator of performance from the last time step) from the environment and makes an action. During this process, the agent tries to optimize (maximize/minimize) the cumulative reward for its action policy. The beauty of this kind of algorithm is the fact that it doesn't need any supervision, since the agent observes the environment and tries to optimize its performance without human intervention. 

RL algorithms come in two categories: policy based algorithms such as Trust Region Policy Optimization (TRPO) \cite{schulman2015trust},   Advantage Actor Critic (A2C) \cite{mnih2016asynchronous},  Proximal Policy Optimization (PPO) \cite{schulman2017proximal} that optimize the policy that maps from states to actions; and value based algorithms such as Q-learning \cite{watkins1992q}, double Q-Learning \cite{van2016deep2} , and soft Q-learning \cite{haarnoja2017reinforcement} that directly maximize the cumulative rewards. While policy  based algorithms have achieved good results and will potentially be applicable for the problem proposed in this paper \cite{belletti2018expert, wu2017flow}, in this paper, we choose deep Q-learning algorithm.  


In the Q-learning approach, the agent learns a 'Q-Value', denoted $Q(s_t, a_t)$, which is a function of observed state $s_t$ and action $a_t$ that outputs the expected cumulative discounted future reward. Here, $t$ denotes the discrete time index. The cumulative discounted future reward is defined as:
$$Q(s_t,a_t) = r_t + \gamma r_{t+1} + \gamma^2 r_{t+2} + \gamma^3 r_{t+3}+...$$

Here, $r_t$ is the reward at each time step, the meaning of which needs to be specified according to the actual problem, and $\gamma<1$ is the discount factor. At every time step, the agent updates its Q function by an update of the Q value:
$$Q(s_t,a_t) \coloneqq Q(s_t,a_t) + \alpha(r_{t+1}+\gamma \max{Q(s_{t+1},a_t)}-Q(s_t,a_t))$$

In most cases, including the traffic control scenarios of interest, due to the complexity of the state space and action space, deep neural networks can be used to approximate the Q function. Instead of updating the Q value, we use the value:
$$Q(s_t,a_t) + \alpha(r_{t+1}+\gamma \max{Q(s_{t+1},a_t)}-Q(s_t,a_t))$$
as the output target of a Q network and do a step of back propagation on the input of $s_t, a_t$.

We utilized two known methods to stabilize the training process \cite{lin1993reinforcement,mnih2013playing}:
\begin{enumerate}
\item Two Q-networks are maintained, a target Q-network and an on-line Q network.  Target Q-network is used to approximate the true Q-values, and the on-line Q-network is back-propagated every step. In the training period, the agent makes decision with the target Q-network, the results from each time instance are used to update the on-line Q-network. At periodic intervals, on-line Q network’s weights are synchronized with the target Q-network. This will keep the agent's decision network relatively stable, instead of changing at every step. 
\item Instead of training after every step an agent has taken, past experience is stored in a memory buffer and training data is sampled from the memory for a certain batch size. This experience replay aims to break the time correlation between samples \cite{mnih2015human}. 
\end{enumerate}

In this paper, we train the traffic lights agents using a Deep Q-network (DQN) \cite{mnih2015human}. With the Q-learning algorithm described above, our work focuses on the definition of agents' actions and the assignment of the states and rewards, which is discussed in the the following subsection \ref{ss:PM}.

\subsection{Parameter Modeling}
\label{ss:PM}

We consider a traffic light controller, which takes reward and state observation from the environment and chooses an action. In this subsection, we introduce our design of actions, rewards, and states for the aforementioned PD-ITSC system problem. 

\subsubsection{Agent action}

In our context, the relevant action of the agent is either to keep the current traffic light phase, or to switch to the next traffic light phase. At every time step, the agent makes an observation and takes action accordingly, achieving intelligent control of traffic.

\subsubsection{Reward}

For traffic optimization problems, the goal is to decrease the average traffic delay of commuters in the network, by using traffic light phasing strategy $\mathcal{S}$. Specifically, find the best traffic light phasing strategy $\mathcal{S}$, such that $t_\mathcal{S} - t_{\min}$ is minimum, where $t_\mathcal{S}$ is the average travel time of commuters in the network, under the traffic control scheme $\mathcal{S}$, and $t_{\min}$ is the physically possible lowest average travel time. Consider traveling the same distance $d$,
$$d = \int_0^{t_{\mathcal{S}}}v_\mathcal{S}(t)dt = t_{\min}v_{\max}$$

Here, $v_{\max}$ is some maximum reasonable speed for the vehicle,  such as the speed limit of the road of interest. $v_{\mathcal{S}(t)}$ denotes the actual vehicle speed under strategy $\mathcal{S}$, at time $t$. Therefore,
$$t_{\min} = \frac{1}{v_{\max}} \int_0^{t_{\mathcal{S}}}v_\mathcal{S}(t)dt$$
$$t_\mathcal{S} - t_{\min} = \int_0^{t_\mathcal{S}}1dt-\frac{1}{v_{\max}} \int_0^{t_{\mathcal{S}}}v_\mathcal{S}(t)dt $$
$$= \frac{1}{v_{\max}}\int_0^{t_\mathcal{S}}v_{\max}-v_{\mathcal{S}}(t)dt$$

Therefore, to get minimum delay $t_\mathcal{S} - t_{\min}$ is equivalent to minimizing at each step $t$, for each vehicle:
\begin{equation}
\frac{1}{v_{\max}}[v_{\max}-v_\mathcal{S}(t)]
\label{eq:loss}
\end{equation}

We note that this is equivalent to maximizing $v_{\mathcal{S}}(t)$, if the $v_{\max}$ on all roads for all cars are the same. If different vehicles have different $v_{\max}$, the reward function is taken as the arithmetic average of the function for all vehicles.

We define the statement in (\ref{eq:loss}) as the penalty of each step. Our goal is to minimize the penalty of each step. Since reinforcement learning tries to maximize the reward (minimize penalty), we define the opposite number of the loss as the reward for the reinforcement learning problem:
\begin{equation}
r_t = -\frac{1}{v_{\max}}[v_{\max}-v_\mathcal{S}(t)]
\label{eq:reward}
\end{equation}


In some cases, especially when the traffic flow is heavy, one can shape the rewards to guide the agent's action, such as avoiding big traffic jams \cite{ng1999policy}. This is certainly an interesting direction for future research.

\subsubsection{State representation}
\label{sss:state_repr}

For optimal decision making, a system should consider as much relevant information about traffic processes as possible. Traditional ITSC system only typically detect simple information such as the presence of vehicles. In PD-ITSC system, only a portion of the vehicles are detected, but it's likely that more specific information about these vehicles such as speed and position are available due to the capabilities of the underlying wireless technologies (discussed in Section \ref{SS:ExampleSystem}). 


RL enables experimentation with many possible choices of inputs and input representations. Further research is required to determine the experimental benefits of each option and that goes beyond the scope of this paper. Based on initial experiments, for the purpose of this paper, we selected a state representation including the
distance to the nearest vehicle at each approach, number of vehicles at each approach, amber phase indicator, current traffic light phase elapsed time and current time, as shown in Table \ref{tab:state_representation}.
\begin{table}[ht]
    \centering
    \caption{details of state representation}
    \label{tab:state_representation}
\begin{tabularx}{0.9\linewidth}{|p{1.7cm}||C|}
\hline
\textbf{Information}&\textbf{Representation}\\
\hline
Detected car count  &Number of detected vehicles in each approach\\
\hline
Distance to nearest detected vehicle & Distance to nearest detected vehicle on each approach; if no detected vehicle, set to lane length (in meters)\\
\hline
Current phase time  & Duration from start of current phase to now (in seconds)\\
\hline
Amber phase  & Indicator of amber phase; 1 if currently in amber phase, otherwise 0 \\
\hline
Current time & Current time of day (hours since midnight), normalized from 0 to 1 (divided by 24)\\
\hline
Current phase & Detected car count and distance to nearest detected vehicle is negated if red, positive if green\\
\hline
\end{tabularx}

\end{table}
Note that current traffic light phase (green or red) is represented by a sign change in the per-lane detected car count and distance rather than by a separate indicator. In initial experiments, we observed slightly faster convergence using this distributed representation (sign representation) than a separate indicator (shown in Figure \ref{fig_train}). We hypothesize that, in combination with Rectified Linear Unit (ReLU) activation, this encoding biases the network to utilize different combinations of neurons for different phases. ReLU units are active if the output is positive and inactive if the output is negative, so our representation may encourage different units to be utilized during different phases, accelerating learning. There are many possible representations and our experimentation with different representations is not exhaustive, but we found that RL was able to handle several different representations with reasonable performance. 

\subsection{System}

\begin{figure}[h]
\centering
\includegraphics[width=3in]{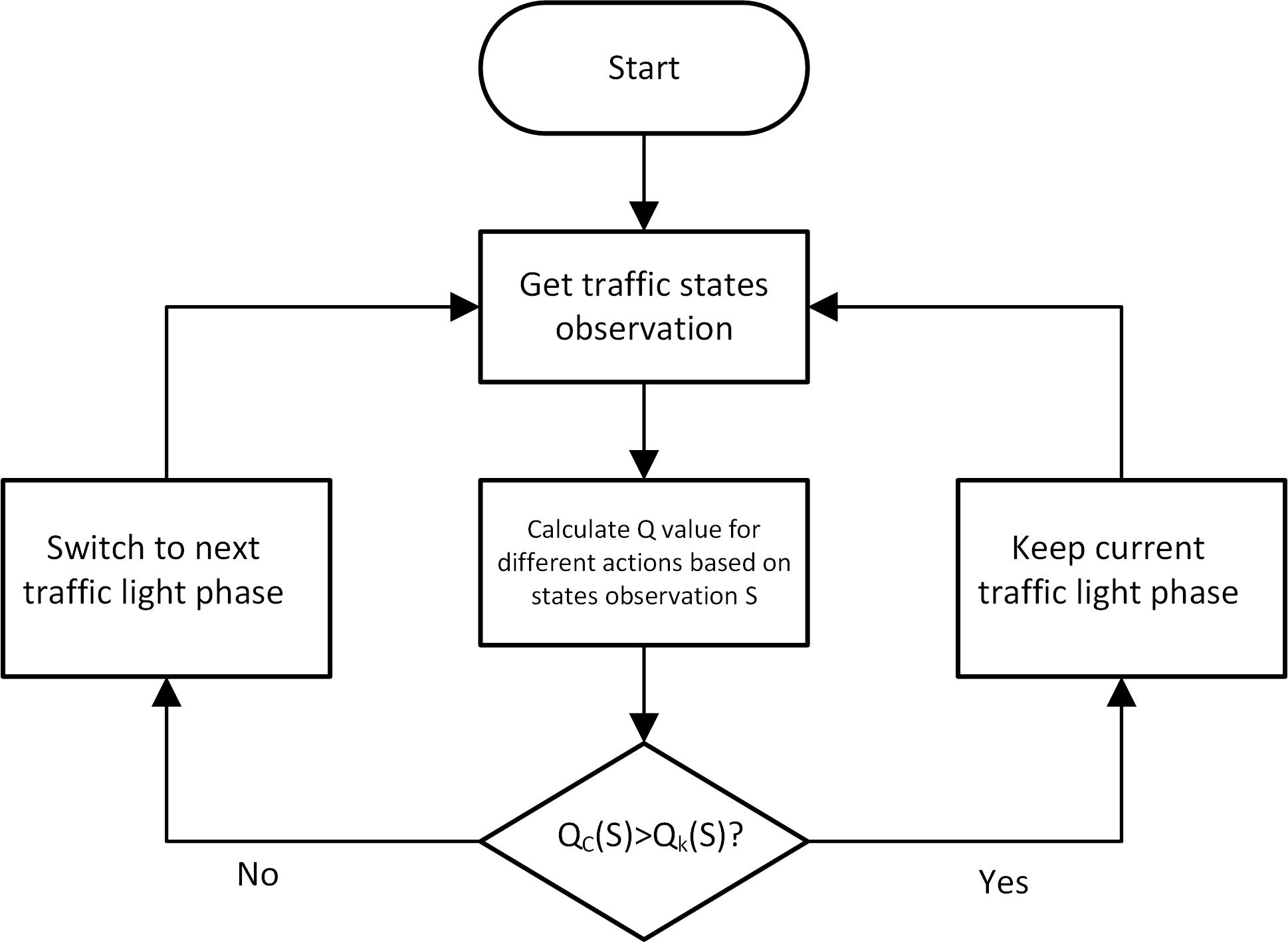}
\caption{Control logic of RL based decision making unit}
\label{fig_control}
\end{figure}

Figure \ref{fig_control} gives a flow chart on how the RL based control unit makes the decisions. As shown in the figure, control unit gets the state representation periodically, calculates the Q-value for all the possible actions and if the action of keeping the current phase has bigger Q-value, it retains the phase; otherwise, switches to the next phase.

Other than the main logic discussed above, a sanity check is performed on the agent: a mandatory maximum and minimum phase. If the current phase duration is less than the minimum phase time, the agent will keep the current phase no matter what action the DQN is choosing; similarly, if phase duration is larger or equal to maximum phase time, the phase will be forced to switch.

\subsection{Implementation}
\label{subsection:implementation}
\begin{figure}[h]
\centering
\includegraphics[width=3in]{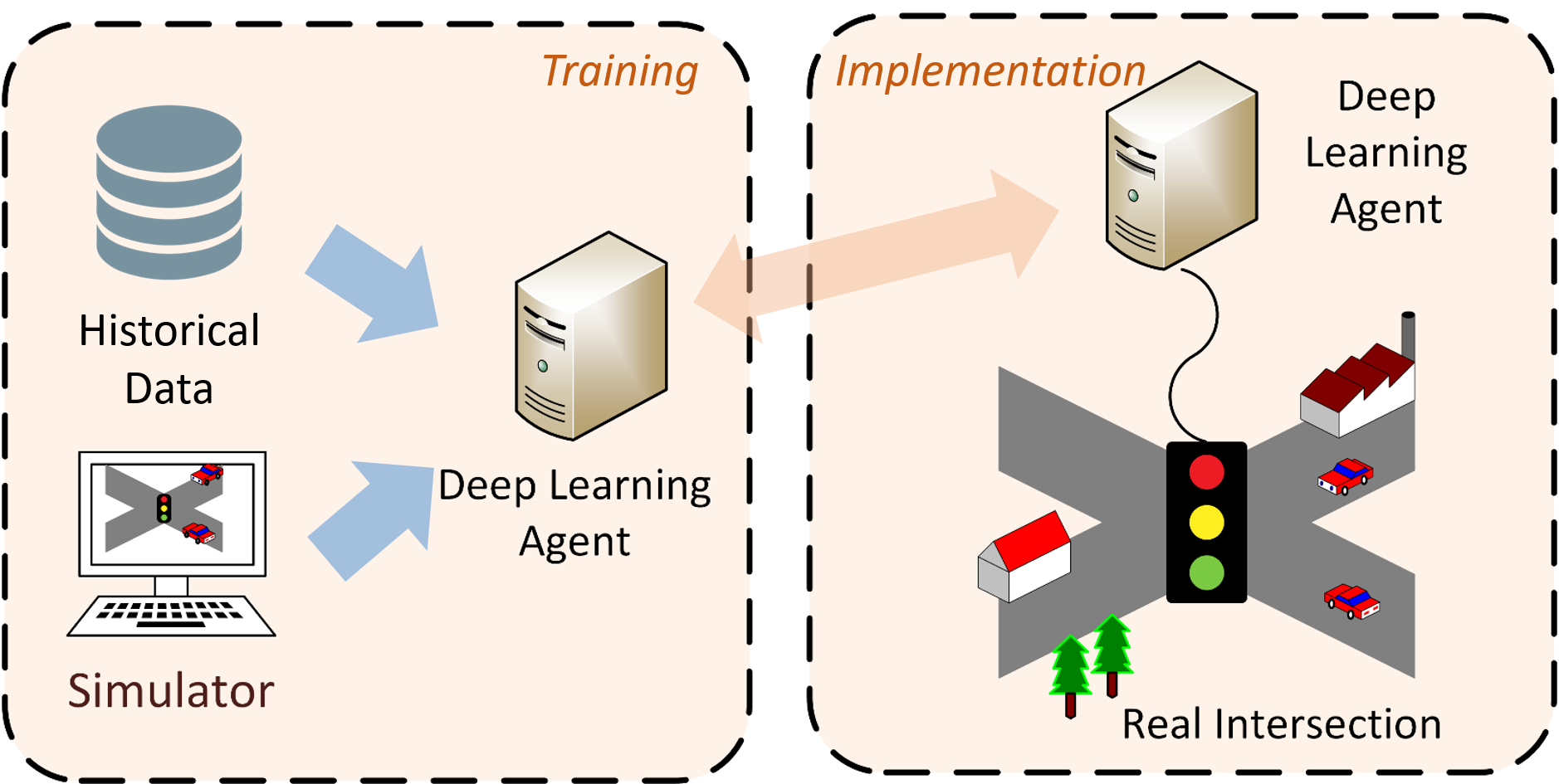}
\caption{The deployment scheme}
\label{fig_deployment}
\end{figure}

In this section, we describe the design of the proposed scheme at the system level. The implementation of the system contains two phases, the training phase and the deployment phase. As shown in Figure \ref{fig_deployment}, the agent is first trained with a simulator, which is then ported to the intersection, connected to the real traffic signal, after which it starts to control the traffic.

\subsubsection{Training phase}
The agent is trained by interacting with a traffic simulator.
The simulator randomly generates vehicle arrivals, then determines whether each vehicle can be detected by drawing from a Bernoulli distribution parameterized by $p$, the detection rate. In the context of DSRC-based vehicle detection systems, the detection rate corresponds to the DSRC penetration rate. The simulator obtains the traffic state $s_t$ and calculates the current reward $r_t$ accordingly, and feeds it to the agent. Using the Q-learning updating formula cited in previous sections, the agent updates itself based on the information from the simulator. Meanwhile, the agent chooses an action $a_t$, and forwards the action to the simulator. The simulator will then update, and change the traffic light phase according to agent’s indication. These steps are done repeatedly until convergence, at which point the agent is trained.


The performance of an agent relies heavily on the quality of the simulator. To obtain similar arrival pattern as the real world, the simulator generates car flow by the historical record of vehicle arrival rate on the same map of the real intersection. To address the variance in car flow in different parts of the day, current time of the day is also specified in the state representation, so that after training the agent is able to adapt to different car flow in different time of the day. Other factors that affect car flow, such as day of the week, could also be parameterized in the state representation.

The goal of training is to have the traffic control scheme achieve the shortest average commute time for all commuters. In the training period, the machine tries different control schemes and eventually converges to an optimal scheme which yields a minimum average commute time. 

\subsubsection{Deployment phase}

In the deployment phase, the software agent is moved to the intersection for controlling the traffic signal. Here, the agent will not update the learned Q-function, but simply control the traffic signal. Namely, the detector will feed the agent's current detected traffic state $s_t$; based on $s_t$, the agent chooses an action based on the trained Q-network and directs the traffic signal to switch/keep phase accordingly. This step is performed in real-time, thus enabling continuous traffic control.

\section{Performance Analysis}
\label{section:performance}
In this section, we give several scenarios of simulations to evaluate various aspects of the performance of the proposed scheme. The simulations are performed with SUMO, a microscopic traffic simulator \cite{krajzewicz2012recent, krauss1997metastable, SUMO2018}. Different scenarios are considered, in order to provide a comprehensive analysis for the proposed scheme.

Qualitatively speaking, we see the performance of the agent reacting to the traffic intelligently from the GUI. It makes reasonable decisions for the arriving vehicles. We demonstrate the performance of the agent after different periods of training in a video available in \cite{RLVideo}.

\begin{figure}[h]
\centering
\includegraphics[width=3in]{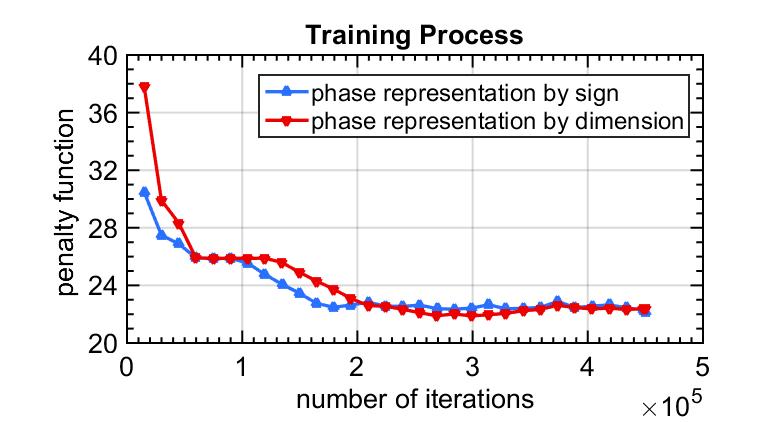}
\caption{Penalty function decreasing with number of iterations in training, the situation shown in the figure is plotted from training with dense car flow at a single intersection}
\label{fig_train}
\end{figure}

Figure \ref{fig_train} shows a typical training process curve. Both phase representations have similar trends, but we do observe that the sign representation has a slightly faster convergence rate in all experiments (see section \ref{sss:state_repr}).

We provide a quantitative analysis in the following subsections.  Though currently there are no analytical results for PD-ITSC system, we can predict what will be observed by considering the following two extreme cases:
\begin{itemize}

    \item When the car flow rate is extremely low, vehicles come to the intersection independently. For detected vehicles, the optimal traffic signal should switch phases on their arrival to yield zero waiting time, for the undetected vehicles, the traffic agent won't be able to do anything. In this case, vehicles can be considered as independent 'particles', and the optimal traffic agent react for each of their arrivals independently. Therefore, we should observe much better performance for the detected vehicles than those undetected vehicles, which corresponds to the cases shown in Figure. \ref{fig_WT_sparse}.
 
    \item When the car flow rate is extremely heavy (at the point of saturation),  the optimal traffic agent should take a completely different strategy, instead of only taking care of the detected vehicles, the agent should be aware of the fact that the detected vehicles are only representatives of the car flow, and react in a way that maximizes the overall waiting time. The waiting time of detected vehicles and undetected vehicles should be similar, because they are of the same car flow. The vehicles here should be considered as 'liquid' instead of 'particles' from the previous case. This can be seen in Figure \ref{fig_WT_dense}.
\end{itemize}


The rest of the section is organized as follows: subsection \ref{ss:DR} evaluates the performance of the system under different detection rates. One should expect different performance for different car flow rates for the reasons mentioned above. 
Subsection\ref{ss:wholeday} gives an estimate on the benefit of the designed agent during different times of the day. Finally, subsection \ref{ss:sensitivity} and \ref{ss:robust} show that when the implementation scenario is slightly different from the training scenario, the performance of the designed agent is still reasonably good.

\subsection{Performance for different detection rates}
\label{ss:DR}
In this subsection, we present performance results under different detection rates, to qualify the performance of a PD-ITSC system as the detection rate increases from 0\% to 100\%. We compare to the performance of a typical pre-timed signal with green phase duration of 24 seconds, shown in dashed lines as a simple reference.


\begin{figure}[h]
\centering
\includegraphics[width=3.5in]{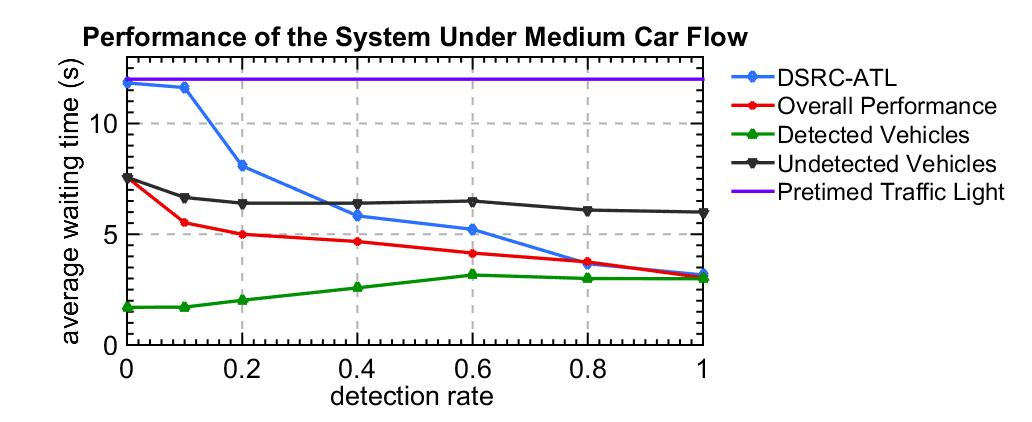}
\caption{Waiting time under different detection rate under medium car flow}
\label{fig_WT_medium}
\end{figure}

Figure \ref{fig_WT_medium} shows a typical trend we obtained in simulations. The figure shows the waiting time of vehicles at a single intersection under the car flow from north, east, south, west to be 0.02 veh/s, 0.1 veh/s, 0.02 veh/s, 0.05 veh/s, respectively, with vehicles arriving as a Poisson process. One can make several interesting observations from this figure. First of all, the system under AI control is much better than the traditional pre-timed traffic signal, even under low detection rate. We can also observe that the overall waiting time (red line) within this system decreases as the detection rate increases. This is intuitive, since as more vehicles are detected, the more information the system has and thus the system is able to optimize the car flow better. 

Additionally, from the figure one can observe that approximately 80\% of the benefit happens in the first 20\% of transition. This finding is quite significant in that we find a transition scheme that asymptotically gets better as the system gradually evolves to a 100\% detection rate, and will be able to receive much of the ultimate benefit during the initial transition.

Another important observation is that during the transition, although the agent is rewarded for optimizing the overall average commute time for both detected and undetected vehicles, the detected vehicles (green line in Figure \ref{fig_WT_medium}) have a lower commute time than undetected vehicles (blue line in Figure \ref{fig_WT_medium}). This provides an interesting 'potential' or 'incentive' to the system, to transition from no vehicles equipped with the IoT device, to all vehicles equipped with the device. Drivers of those vehicles not yet equipped with the device now have a good reason and strong incentive to install one.

Here, we also compare with our previous designed system known as DSRC-ATL \cite{zhang2018increasing}, which is an algorithm designed for dealing with partial detection under sparse to medium car flow. We see that though the algorithms exhibit similar trends, RL agents have better performance during the whole transition from 0 to 1 detection rate. 
\begin{figure}[ht]
\centering
\subfloat[Performance under dense flow]{\includegraphics[width=3.5in]{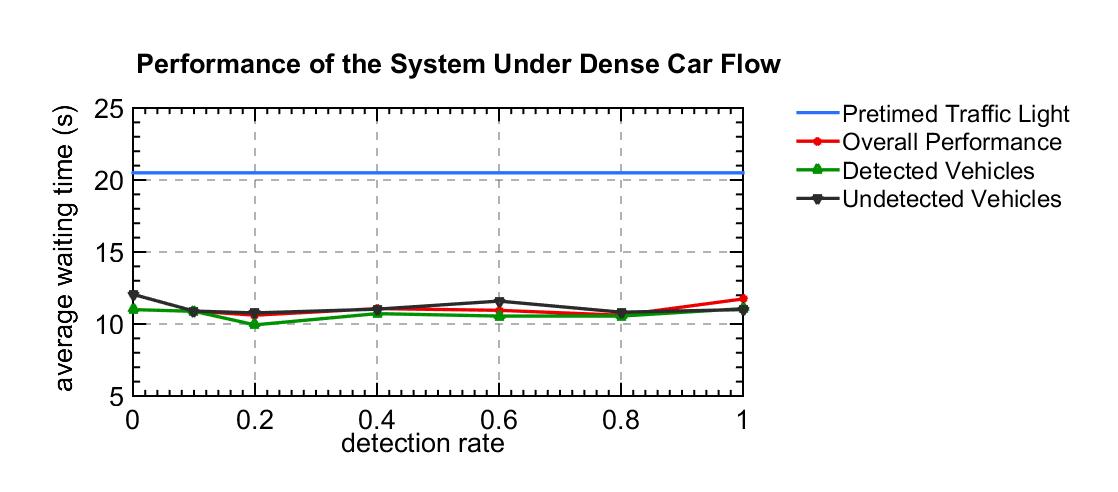}%
\label{fig_WT_dense}}
\hfil
\subfloat[Performance under sparse flow]{\includegraphics[width=3.5in]{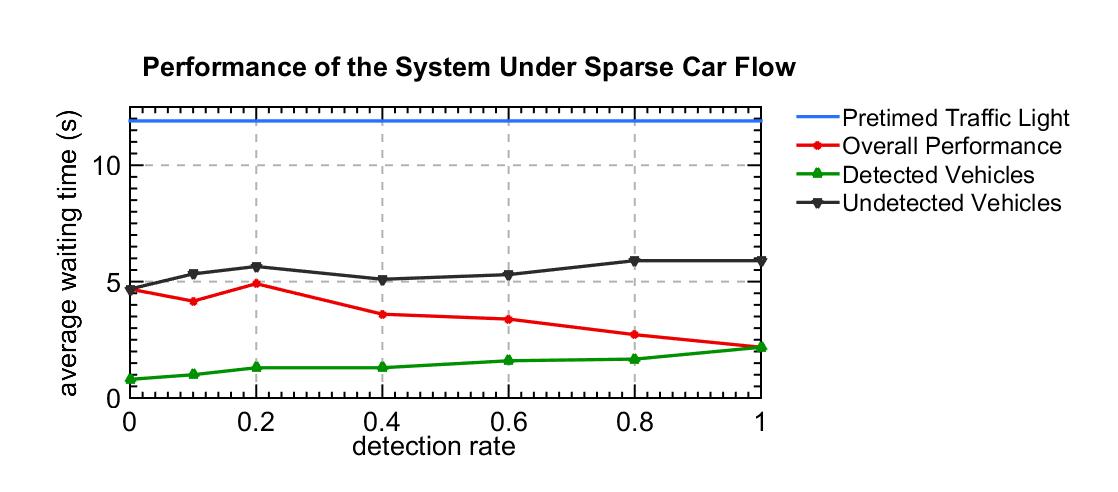}%
\label{fig_WT_sparse}}
\caption{Waiting time under different detection rate under dense and sparse car flow}
\label{fig_WT_dense_sparse}
\end{figure}

Figure \ref{fig_WT_dense_sparse} shows the performance under the other two cases: when the car flow is very sparse (0.02 veh/s at each lane) or very dense (0.5 veh/s at each lane). For the sparse situation in Figure \ref{fig_WT_sparse}, the trend is similar to the medium flow case shown in Figure \ref{fig_WT_medium}.  

One can see from Figure \ref{fig_WT_dense} that under the dense situation, the curve becomes quite flat. This is because when car flow is high, detecting individual vehicles become less important. When many cars arrive at the intersection, car flow has 'liquid' qualities, as opposed to 'particle' qualities in the previous two situations. The trained RL agent is able to seamlessly transition from a 'particle arrival' optimization agent which handles random arrivals to a 'liquid arrival' optimization agent which handles macroscopic flow. This result shows that RL is able to capture the main factors that affect traffic system's performance and performs differently under different car arrival rates. Hence, RL provides a much desired adaptive behavior.

\subsection{Performance of a whole day}
\label{ss:wholeday}
\begin{figure}[hbt]
\centering
\includegraphics[width=3in]{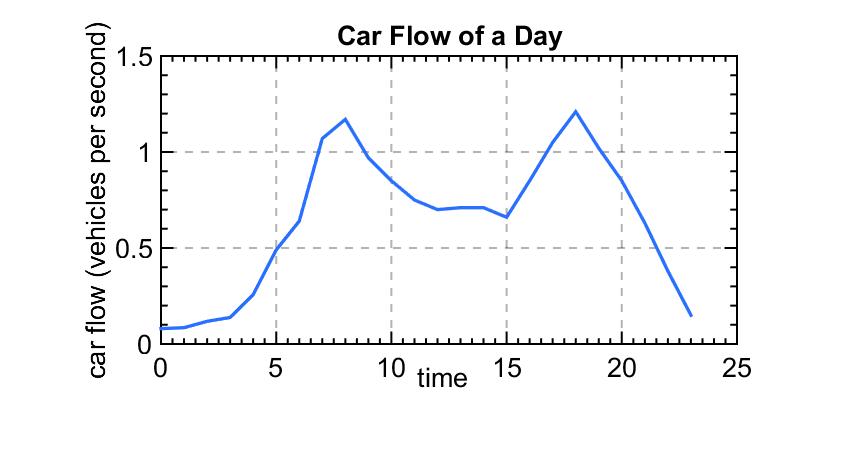}
\caption{Typical car flow in a day}
\label{fig_carFlow}
\end{figure}

Section \ref{ss:DR} examines the effect of flow rate on system performance. Since the car flow differs at different times of the day, we simulate an entire day of traffic. To generate realistic car flow of a day, we refer to the whole day car flow reported in \cite{flow}. To adapt the reported arrival rate to the simulation system, we multiply the car flow in \cite{flow} with a factor so that the peak volume matches the saturation flow rate of the simulated roads. Figure \ref{fig_carFlow} shows the car flow rate we used for the simulation, the car flow reach peak on 8 am in the morning and 6 pm in the afternoon of 1.2 vehicles/s, the car flow of the regular hours is around 0.7 vehicles/s. It is worth mentioning that the car flow of different intersections in the real world might be very different, so the result presented here is just an example of what the performance looks like under a typical traffic volume of a whole day.

\begin{figure}[h]
\centering
\includegraphics[width=3.5in]{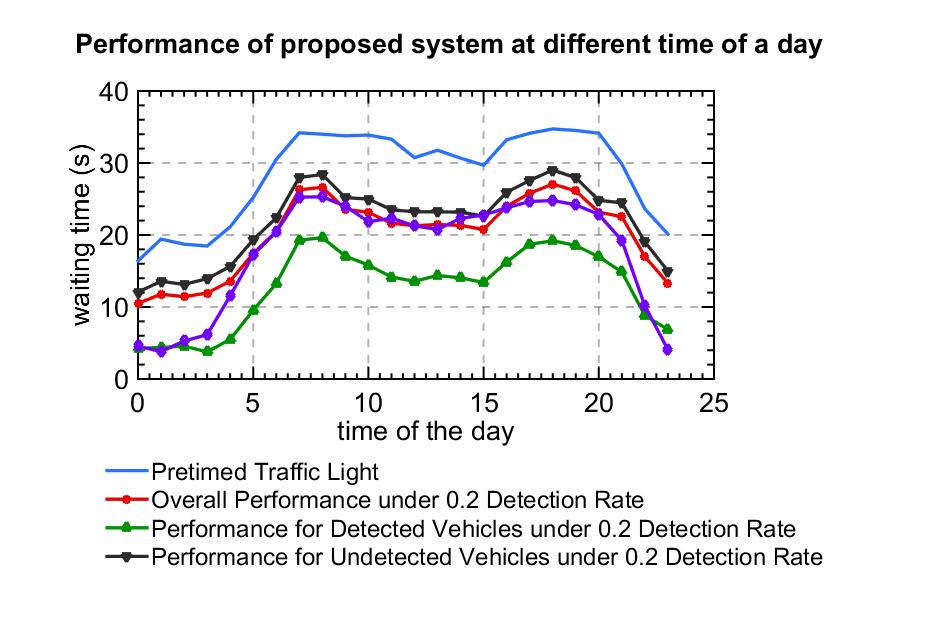}
\caption{Expected Performance by Time}
\label{fig_daytime_performance}
\end{figure}

Figure \ref{fig_daytime_performance} shows the performance of different vehicles in a whole day. One can observe from this figure that the performance of 20\% detection rate (red line) is very close to the performance of 100\% detection rate (green line), at most times of the day (from 5am to 9pm). During rush hours, the system with 100\% detection rate is almost the same as the system with 20\% detection rate.  Though a traffic system under 100\% detection rate performs visibly better at midnight, the performance at that time is not as critical as the performance during the busier daytime. This result indicates that by detecting 20\% of vehicles, we can perform almost the same as detecting all vehicles. But those detectable vehicles (yellow lines) will have a benefit against those undetectable vehicles (dash line).

These results confirm intuition. With a large volume of cars, a low detection rate should still provide a relatively low-variance estimate of traffic flow. If there are few cars and a low detection rate, the estimate of traffic flow can have very high-variance. Late at night with only a single detected car, an ITSC system can give that car a green immediately, which would not be possible with an undetected car.

\subsection{Sensitivity Analysis}
\label{ss:sensitivity}
The results obtained above used agents trained and evaluated under the same environmental parameters, since traffic patterns only fluctuate slightly from day to day.

Below, we evaluate the sensitivity of the agents to two environmental parameters: the car flow and the detection rate.

\subsubsection{Sensitivity to car flow}
\begin{figure}[!h]
\centering
\subfloat[Sensitivity of agent trained under 0.1 veh/s flow rate]{\includegraphics[width=.5\linewidth]{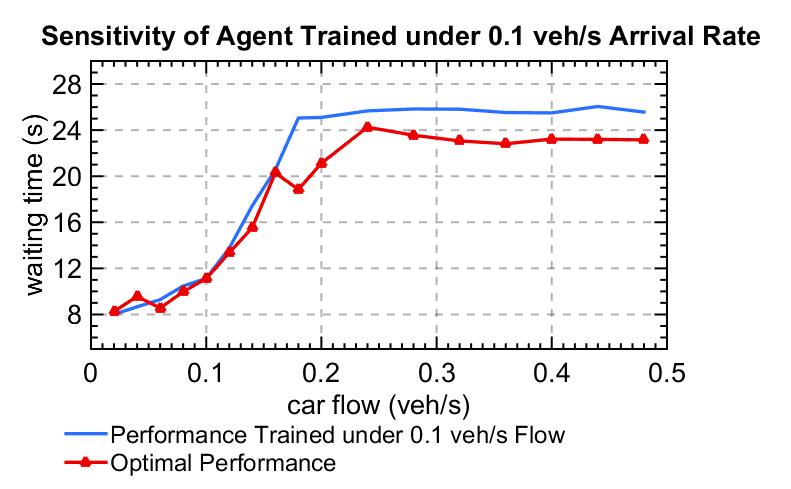}%
\label{fig_flow_0_1}}
\hfil
\subfloat[Sensitivity of agent trained under 0.5 veh/s flow rate]{\includegraphics[width=.5\linewidth]{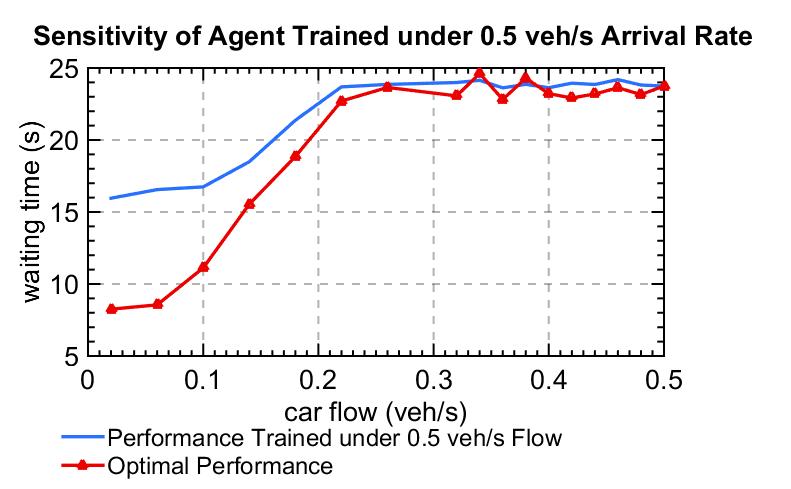}%
\label{fig_flow_0_5}}
\caption{Sensitivity analysis of flow rate}
\label{fig_flow_sensitivity}
\end{figure}

Figure \ref{fig_flow_sensitivity} shows the agents' sensitivity to car flow. Figure \ref{fig_flow_0_1} shows the performance of an agent trained under 0.1 veh/s car flow, operating at different flow rates. Figure \ref{fig_flow_0_5} shows the sensitivity of an agent trained under 0.5 veh/s car flow. The blue curve in the figure is the trained agent's performance, while the red one is the performance of the optimal agent (the agent trained under that situation and tested under that situation). Both agents perform well over a range of flow rates. The agent trained under 0.1 veh/s flow can handle flow rates from 0 to 0.15 at near-optimal levels. At higher flow rates, it still performs reasonably well. The agent trained on 0.5 veh/s flow will perform reasonably from 0.25 veh/s to 0.5 veh/s, but under 0.25 veh/s, 
the agent will start to perform substantially worse than the optimal agent. 
Since traffic patterns are not expected to heavily fluctuate, these results give a strong indication that the agent trained by the data will be able to adapt to the environment even when the trained situation is slightly different.

\subsubsection{Sensitivity to detection rate}
In most situations, the detection rate can only be approximately measured. It is likely that an agent trained under one detection rate needs to operate under a slightly different detection rate, so we test the sensitivity of agents to detection rates.

\begin{figure}[!h]
\centering
\subfloat[Sensitivity of agent trained under 0.2 detection rate]{\includegraphics[width=.5\linewidth]{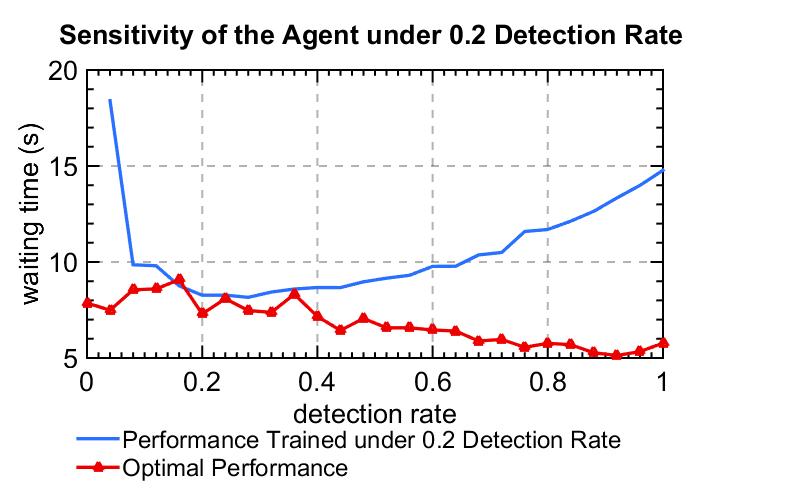}%
\label{fig_pen_0_2}}
\hfil
\subfloat[Sensitivity of agent trained under 0.8 detection rate]{\includegraphics[width=.5\linewidth]{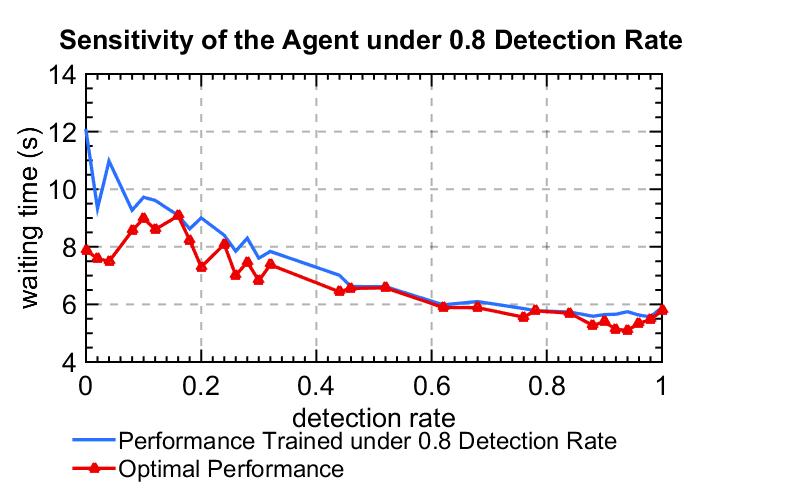}%
\label{fig_pen_0_8}}
\caption{Sensitivity analysis of detection rate}
\label{fig_pen_sensitivity}
\end{figure}

Figure \ref{fig_pen_sensitivity} shows the sensitivity of two cases. Figure \ref{fig_pen_0_2} shows the sensitivity of low detection rate (0.2), figure \ref{fig_pen_0_8} shows the sensitivity under high detection rate (0.8). 

We observe that the agent trained under 0.2 detection rate performs at an optimal level from 0.1 to 0.4 detection rate. The sensitivity upward is better than downward. This indicates that at early deployment of this system, it's better to under-estimate detection rate, since the agent's performance is more stable for the higher detection rate.


Figure \ref{fig_pen_0_8} shows the sensitivity of the agent trained under high detection rate (0.8). We can see that the performance of this agent is at optimal level when detection rate is from 0.5 to 1. Though the sensitivity performance for an agent under low detection rate is different than the sensitivity under high detection rate, for both cases, the agent shows a level of stability, which means that as long as the detection rate used for training is not too different from the actual detection rate, the performance of the agent will not be affected a lot.

\subsection{Robustness between training and deployment scenario}
\label{ss:robust}
There are many differences between the training and the actual deployment scenario, as the simulator, though quite sophisticated, will never able to take all the factors in the real scenario into account. This simulation aims to evaluate and verify that those minor factors, such as stop-and-go vehicles, arrival patterns and other factors won't affect the system in a major way. We choose a newly published realistic scenario known as Luxembourg SUMO Traffic (LuST) \cite{codeca2017luxembourg}. The scenario is generated on the real map of Luxembourg, the activity of vehicles are generated according to the demographic data published by the government. The authors of this scenario compared the generated traffic with a data set collected between March and April 2015 in Luxembourg, which contains 6,000,000 floating vehicles sample and achieved similar speed distributions, hence the LuST scenario has a high degree of realism.

In our simulation, we don't directly train the traffic light on the scenario; instead, we use this scenario as ground truth to evaluate the trained traffic light. The simulation steps we performed are as follows:
\begin{enumerate}
    \item Choose a certain intersection from LuST with high rate of car flow (intersection -12408)
    \item Measure the hourly traffic volume of that intersection
    \item Build a simple intersection in a separate simulator and train a traffic agent with car flow generated by the new simulator, according to the hourly traffic volume measured in step 2. 
    \item Train an agent on the simplified scenario we built in step 3.
    \item After training, we evaluate the performance on the original LuST scenario, by substituting the traffic agent of that intersection to the new traffic agent we trained.
\end{enumerate}

It is worth mentioning here that this simulation follows the steps of actual implementation in real world (described in section IV-D), so the performance here can be considered  as a reference for the performance of actual deployment when the simulator and real world have major differences in details.

Other than the difference in the map and car flow, there are more differences between training and evaluation, the scenario used for evaluation is rich in details. In Table \ref{tab:robust}, we list all the differences between the Lust scenario (for evaluation) and the simulator used for training. 

\begin{table}[ht]
    \centering
    \caption{Deference in training and evaluation scenario}
    \label{tab:robust}
\begin{tabularx}{0.95\linewidth}{|p{1.4cm}||C|C|}
\hline

&\textbf{training} & \textbf{Evaluation (LuST)}\\
\hline
Map topology & Simple straight street intersection & Real world map\\
\hline
Street length & 125m for each approach & Different length for each approach\\
\hline
Car arrival pattern & Poisson & Bulk arrival when vehicle go through intersections\\
\hline
Car speed & Constant & Gaussian mixture distribution\\
\hline
Stop-and-go & No stop-and-go vehicles& Bus stops\\
\hline
U-turn vehicles & No U-turn & A small proportion of U-turn\\
\hline
Location where vehicle generated & End of the road & Anywhere of the road \\
\hline
Location of destination & End of the road & anywhere of the road, some might not even go through the intersection \\
\hline
Buses & No buses & Regular buses arrival with a bus stop close to the intersection\\
\hline
Vehicle passing & Almost no passing due to constant speed & Some vehicle passing due to the randomness of the speed\\
\hline
\end{tabularx}
    
\end{table}
Notice that the simulator is sophisticated enough to take all the factors listed in the table into account. Here we intentionally introduce differences between training and evaluation. This is a judicious choice on our part. Our goal is to give a reasonable estimate of the performance in the real-world implementation where the simulation scenario is slightly different than the real-world scenario.

We choose three different times of the day to present the results:
\begin{enumerate}
    \item \emph{Midnight}: 2 AM in the morning, in this case, the car flow at intersection is sparse
    \item \emph{Rush-hours}: 8 AM in the morning, this is a situation where car flow is dense
    \item 	\emph{Regular hours}: 2 PM in the afternoon, this is the situation during regular hours, the car flow is in between of midnight car flow and rush hours car flow (medium car flow).
\end{enumerate}

\begin{figure}[hbt]
\centering
\subfloat[Performance of traffic agent in 2 am]{\includegraphics[width=.9\linewidth]{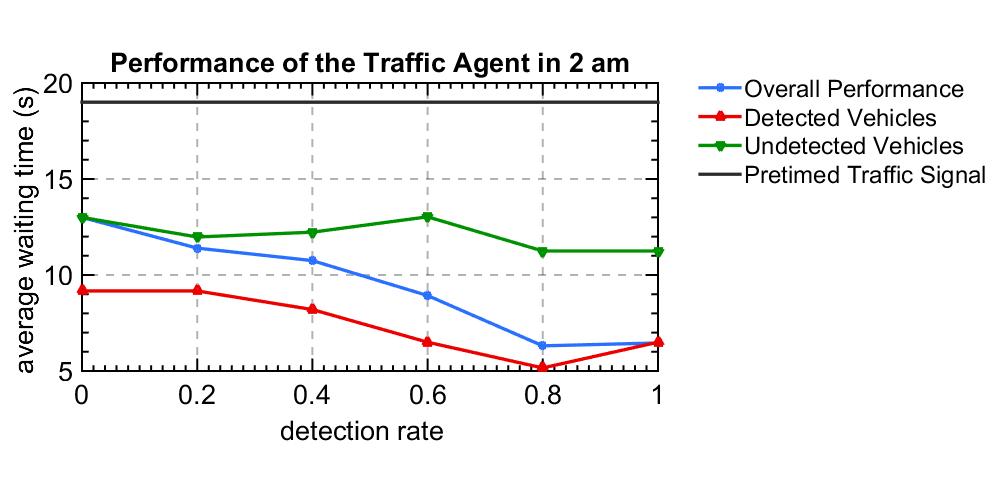}%
\label{fig_eval_2}}
\hfil
\subfloat[Performance of traffic agent in 8 am]{\includegraphics[width=.9\linewidth]{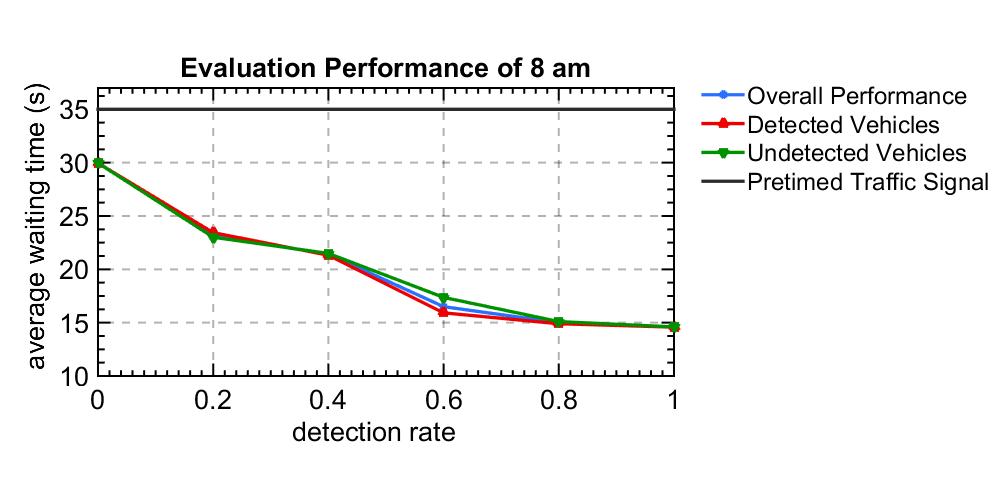}%
\label{fig_eval_8}}
\hfil
\subfloat[Performance of traffic agent in 2 pm]{\includegraphics[width=.9\linewidth]{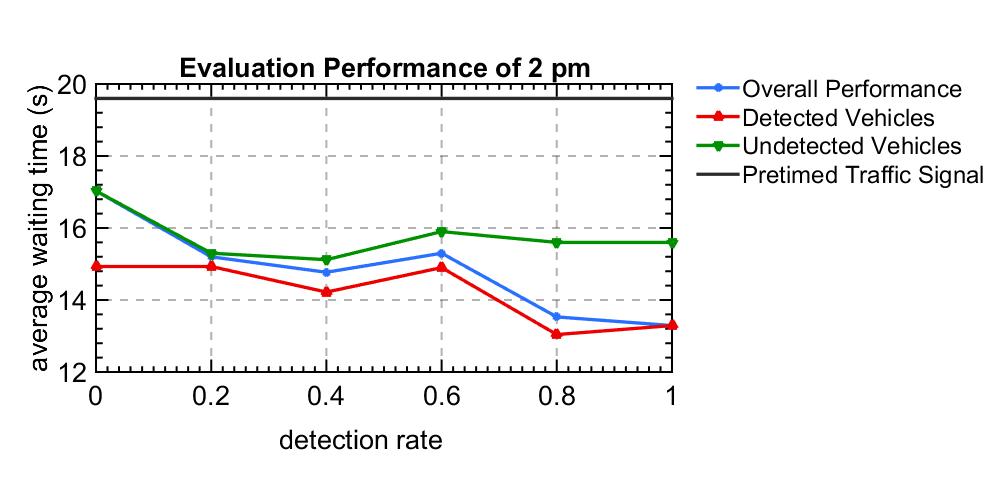}%
\label{fig_eval_14}}
\caption{Performance of the agent in LuST scenario}
\label{fig_robust}
\end{figure}

Figure \ref{fig_robust} shows the performance of the agent in the LuST scenario. We can clearly see that even though the evaluated situation is quite different from the training situation, we still observe: the performance improves asymptotically as the detection rate grows, which exhibits the same trend as we observed in \ref{ss:DR}. 
\section{Discussion}
\label{ss:discussion}
As the simulation results show, while all vehicles will experience a shorter waiting time under an RL-based traffic controller, detected vehicles will have a shorter commute time than undetected vehicles. This property makes it possible for hardware manufacturers, software companies, and vehicle manufacturers to help push forward the proposed scheme, other than the Department of Transportation (DoT) alone, for the simple reason that all of them can profit from this system. For example, it would be valuable for a certain navigation app to advertise that their customers can save 30\% on commute time.

Therefore, we view this technology as a new generation of Intelligent Transportation Systems, as it inherently comes with a lucrative commercial business model. The burden of spreading the penetration rate in this system is distributed to a lot of companies, as opposed to the traditional ITSC systems which put all the burden on the DoT alone. This makes it financially feasible to have the system installed on most of the intersections in a city, as opposed to the current situation where only a small proportion of intersections are installed with ITSC.

The mechanism of the system solution described will also make it possible to have dynamic pricing. Dynamic pricing refers to reserving certain roads during rush hours exclusively for paid users. This method has been scuttled by public or political opposition and only a few cities have implemented dynamic pricing \cite{de2011traffic, schaller2010new}. Those few successful examples, however, cannot be easily copied or adapted to other cities, as the method depends hugely on road topologies.. In our solution, we can accomplish dynamic pricing in a more intelligent way, by simply consider vehicle detection as a service. 
Compared to existing solutions, this service will not require to reserve roads, making the scheme flexible and easy to implement. The user will also be able to choose to pay for a prioritized signal phase whenever they are in a hurry. 

Further research is needed to make this AI-based Intelligent Traffic Control System more practical. First of all, the system currently needs to be fully trained in a simulator; under the partial observation setup, the system will not be able to observe the reward, hence, it won't be able to do any incremental training after deployment. Clearly, this is a drawback or shortcoming of the proposed system. Some solutions to this problem are reported in a follow-up paper \cite{zhang2019partially}. Another future direction would be to further develop the system to achieve multi-agent coordination so that, with the help of DSRC radios (or other forms of communications), traffic lights will be able to communicate with each other. Clearly, designing such a system will significantly improve the performance of PD-ITSC system. Further research is also required to investigate whether the RL agent will be able to pick up the drivers' behavior accurately at each intersection \cite{noyce2000traffic, tang2007comparative, gates2010dilemma, rittger2015driving, li2016predicting}.

\section{Conclusion}
\label{section:conclusion}
In this paper, we have proposed reinforcement learning, specifically deep Q-learning, for traffic control with partial detection of vehicles. The results of our study show that reinforcement learning is a promising new approach to optimizing traffic control problems under partial detection scenarios, such as traffic control systems using DSRC technology. This is a very promising outcome that is highly desirable since the industry forecasts on DSRC penetration process seems gradual as opposed to abrupt. 

The numerical results on sparse, medium, and dense arrival rates suggest that reinforcement learning is able to handle all kinds of traffic flow. Although the optimization of traffic on sparse arrival and dense arrival are, in general, very different, results show that reinforcement learning is able to leverage the
’particle’ property of the vehicle flow, as well as the ’liquid’ property, thus providing a very powerful overall optimization scheme.



%

\section*{Acknowledgment}

The authors would like to thank to Dr. Hanxiao Liu from Language Technology Institute, Carnegie Mellon University for informative discussions and a lot of suggestions to the methods reported in the paper. The authors would also like to thank Dr. Laurent Gallo from Eurecom, France and Mr. Manuel E. Diaz-Granados of Yahoo, US, for the initial attempt to solve this problem in 2016.

\ifCLASSOPTIONcaptionsoff
  \newpage
\fi



%
\bibliographystyle{IEEEtran}
\vspace{-0.05in}
\bibliography{IEEEabrv,reference}

%

\begin{IEEEbiography}[{\includegraphics[width=1in,height=1.25in,clip,keepaspectratio]{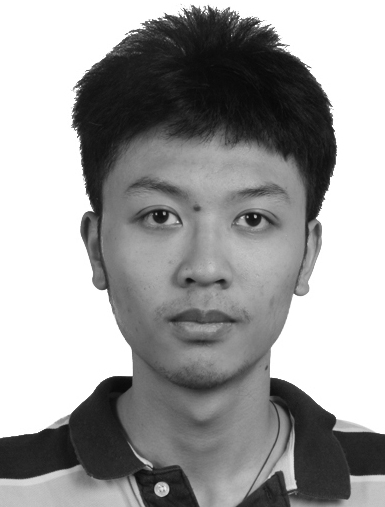}}]{Rusheng Zhang}
was born in Chengdu, China in 1990. He received the B.E. degree in micro electrical mechanical system and second B.E. degree in Applied Mathematics from Tsinghua University, Beijing, in 2013, and the M.S. degree in electrical and computer engineering from Carnegie Mellon University, in 2015. He is a Ph.D. candidate at Carnegie Mellon University. His research areas include vehicular networks, intelligent transportation systems, wireless computer networks, artificial intelligence and intra vehicular sensor networks.
\end{IEEEbiography}
\vskip -2\baselineskip plus -1fil
\begin{IEEEbiography}[{\includegraphics[width=1in,height=1.25in,clip,keepaspectratio]{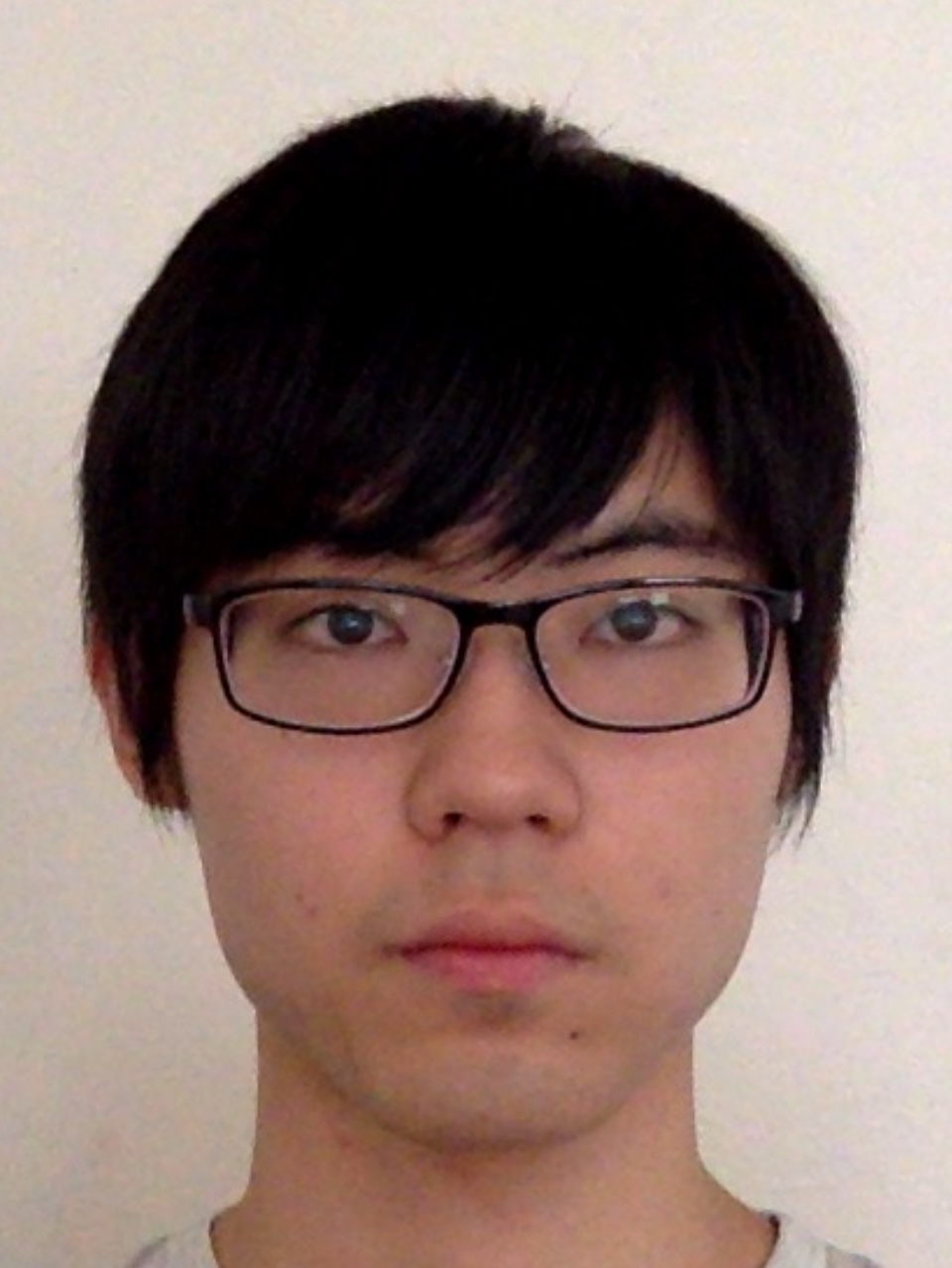}}]{Akihiro Ishikawa}
was an MS student in the Electrical and Computer Engineering Department of Carnegie Mellon University until he received his MS degree in 2017. His research interests include vehicular networks, wireless networks, and artificial intelligence. 
\end{IEEEbiography}
\vskip -2\baselineskip plus -1fil
\begin{IEEEbiography}[{\includegraphics[width=1in,height=1.25in,clip,keepaspectratio]{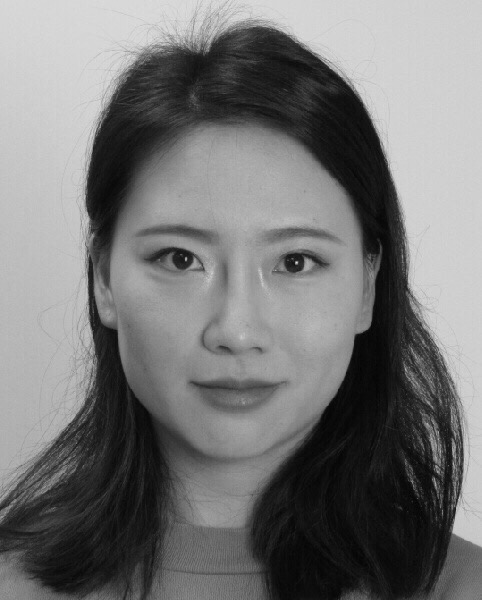}}]{Wenli Wang}
has obtained an M.S. degree in the Electrical and Computer Engineering Department of Carnegie Mellon University in 2018. Prior to Carnegie Mellon University, she received B.S. in Statistics and B.A. in Fine Arts from University of California, Los Angeles in 2016. Her research interests include machine learning and it applications in wireless networks and computer vision. 
\end{IEEEbiography}
\vskip -2\baselineskip plus -1fil
\begin{IEEEbiography}[{\includegraphics[width=1in,height=1.25in,clip,keepaspectratio]{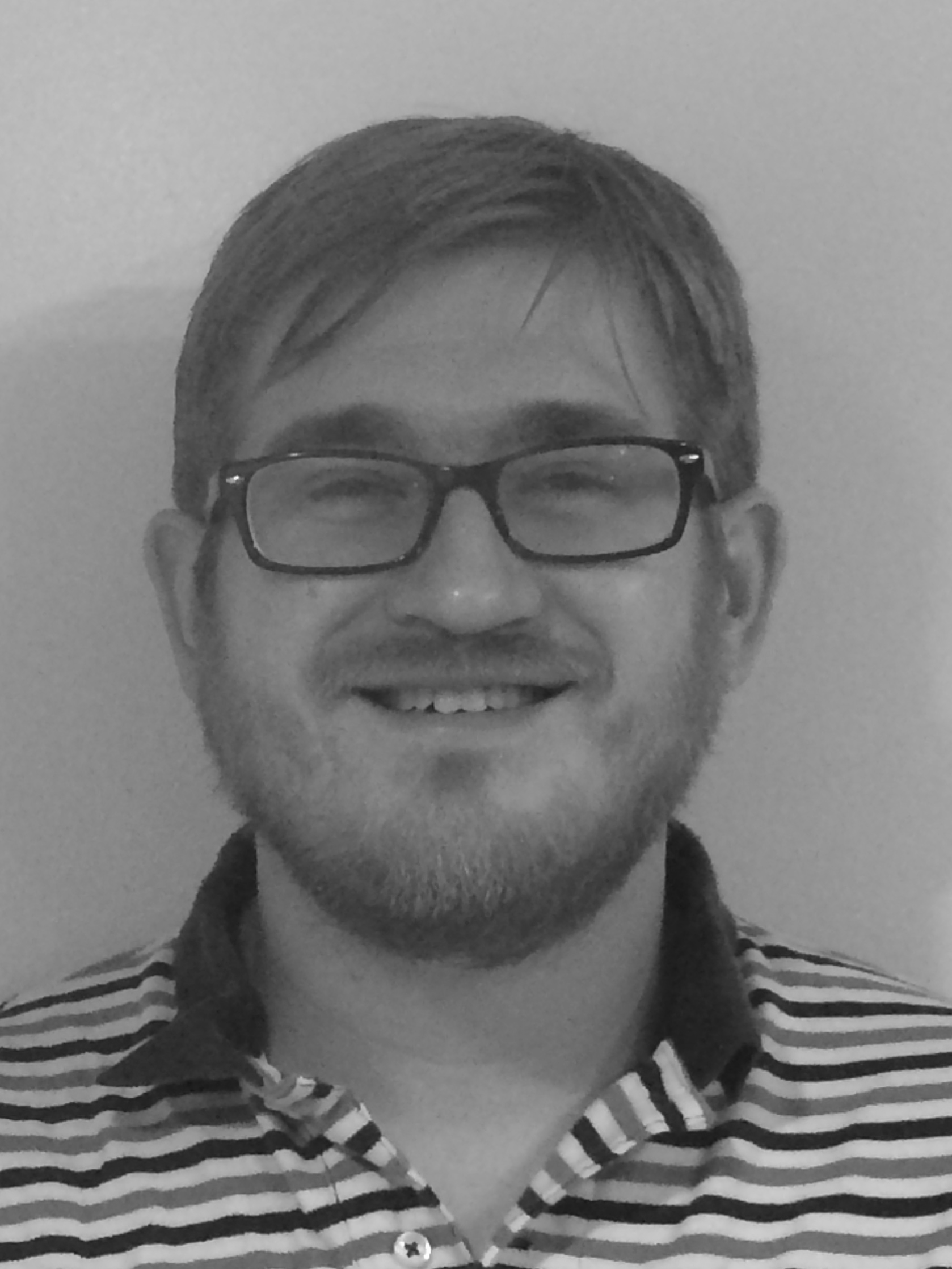}}]{Benjamin Striner} is a master's student in the Machine Learning Department at Carnegie Mellon University. Previously, he was a patent expert witness and engineer, especially in wireless communications. He received a B.A. in neuroscience and psychology from Oberlin College in 2005. Research interests include reinforcement learning, generative networks, and better understandability and explainability in machine learning. 
\end{IEEEbiography}
\vskip -2\baselineskip plus -1fil
\begin{IEEEbiography}[{\includegraphics[width=1in,height=1.25in,clip,keepaspectratio]{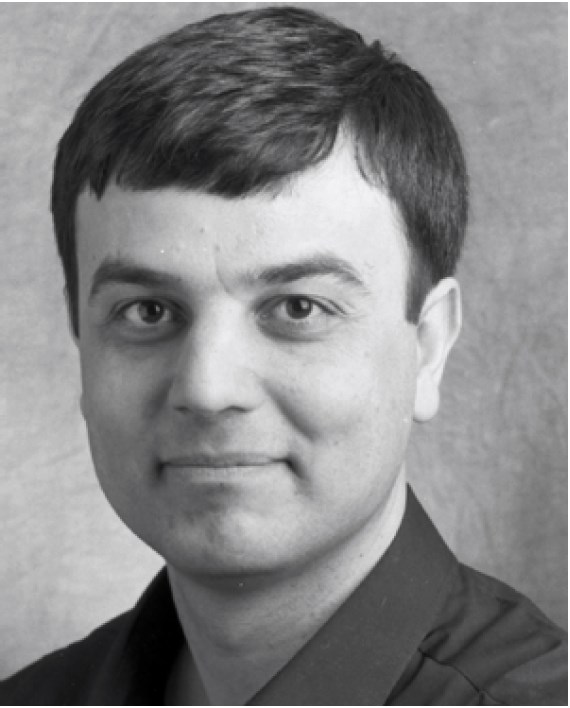}}]{Ozan Tonguz}
is a tenured full professor in the Electrical and Computer Engineering Department of Carnegie Mellon University (CMU). He currently leads substantial research efforts at CMU in the broad areas of telecommunications and networking. He has published about 300 research papers in IEEE journals and conference proceedings in the areas of wireless networking, optical communications, and computer networks. He is the author (with G. Ferrari) of the book \emph{Ad Hoc Wireless Networks: A Communication-Theoretic Perspective (Wiley, 2006)}. He is the inventor of 15 issued or pending patents (12 US patents and 3 international patents). In December 2010, he founded the CMU startup known as Virtual Traffic Lights, LLC, which specializes in providing solutions to acute transportation problems using vehicle-to-vehicle (V2V) and vehicle-to-infrastructure (V2I) communications paradigms. His current research interests include vehicular networks, wireless ad hoc networks, sensor networks, self-organizing networks, artificial intelligence (AI), statistical machine learning, smart grid, bioinformatics, and security. He currently serves or has served as a consultant or expert for several companies, major law firms, and government agencies in the United States, Europe, and Asia.
\end{IEEEbiography}







\end{document}